\documentclass[10pt,twocolumn,letterpaper]{article}

\usepackage{cvpr}              

\usepackage{graphicx}
\usepackage{amsmath}
\usepackage{amssymb}
\usepackage{booktabs}
\usepackage{bm}

\usepackage[utf8]{inputenc} 
\usepackage[T1]{fontenc}    
\usepackage{url}            
\usepackage{amsfonts}       
\usepackage{nicefrac}       
\usepackage{microtype}      
\usepackage{xcolor}         
\usepackage{multirow}
\usepackage[linesnumbered,ruled,vlined]{algorithm2e}

\usepackage{enumitem}
\usepackage{array}
\usepackage{makecell}
\usepackage{adjustbox}
\usepackage{hhline}
\usepackage{tabularx}
\usepackage{floatrow}
\usepackage{csquotes}
\newfloatcommand{capbtabbox}{table}[][\FBwidth]

\definecolor{commentcolor}{RGB}{110,154,155}   
\newcommand{\bb}[1]{\boldsymbol{\mathbf{#1}}}

\usepackage[pagebackref,breaklinks,colorlinks]{hyperref}

\usepackage[capitalize]{cleveref}
\crefname{section}{Sec.}{Secs.}
\Crefname{section}{Section}{Sections}
\Crefname{table}{Table}{Tables}
\crefname{table}{Tab.}{Tabs.}

\def\cvprPaperID{9731} 
\def\confName{CVPR}
\def\confYear{2023}

\begin{document}

\title{PanoHead: Geometry-Aware 3D Full-Head Synthesis in 360$^{\circ}$}



\author{Sizhe An$^{1,2}$ \quad  Hongyi Xu$^{1}$ \quad Yichun Shi$^{1}$ \quad Guoxian Song$^1$ \quad Umit Y. Ogras$^{2}$ \quad Linjie Luo$^1$\\
$^1$ByteDance Inc. \quad\quad $^2$University of Wisconsin-Madison\\
}


\twocolumn[{
\renewcommand\twocolumn[1][]{#1}
\maketitle
\begin{center}
    \captionsetup{type=figure}
    \includegraphics[width=1.0\textwidth]{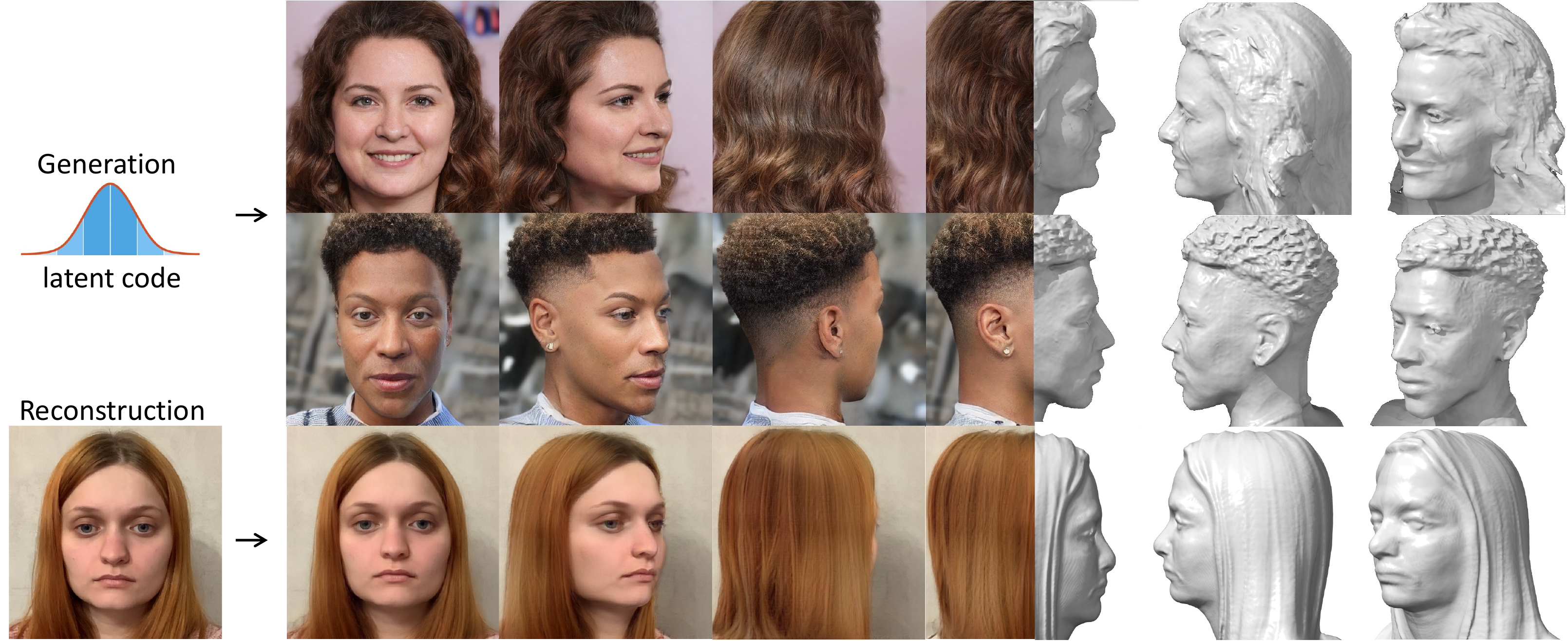}
    \captionof{figure}{Our PanoHead enables $360^\circ$ view-consistent photo-realistic full-head image synthesis with high-fidelity geometry, enabling authentic 3D portraits creation from a single-view image.}
    \label{fig:teaser}
\end{center}
}]


\begin{abstract}
Synthesis and reconstruction of 3D human head has gained increasing interests in computer vision and computer graphics recently. Existing state-of-the-art 3D generative adversarial networks (GANs) for 3D human head synthesis are either limited to near-frontal views or hard to preserve 3D consistency in large view angles. We propose PanoHead, the first 3D-aware generative model that enables high-quality view-consistent image synthesis of full heads in $360^\circ$ with diverse appearance and detailed geometry using only in-the-wild unstructured images for training. At its core, we lift up the representation power of recent 3D GANs and bridge the data alignment gap when training from in-the-wild images with widely distributed views. Specifically, we propose a novel two-stage self-adaptive image alignment for robust 3D GAN training. We further introduce a tri-grid neural volume representation that effectively addresses front-face and back-head feature entanglement rooted in the widely-adopted tri-plane formulation. 
Our method instills prior knowledge of 2D image segmentation in adversarial learning of 3D neural scene structures, enabling compositable head synthesis in diverse backgrounds. Benefiting from these designs, our method significantly outperforms previous 3D GANs, generating high-quality 3D heads with accurate geometry and diverse appearances, even with long wavy and afro hairstyles, renderable from arbitrary poses. Furthermore, we show that our system can reconstruct full 3D heads from single input images for personalized realistic 3D avatars.


\end{abstract}

\newcommand\blfootnote[1]{%
  \begingroup
  \renewcommand\thefootnote{}\footnote{#1}%
  \addtocounter{footnote}{-1}%
  \endgroup
}

\blfootnote{Project page: \url{https://sizhean.github.io/panohead}}


\vspace{-0.1in}
\section{Introduction}

Photo-realistic portrait image synthesis has been a continuous focus in computer vision and graphics, with a wide range of downstream applications in digital avatars, telepresence, immersive gaming, and many others. Recent advances in Generative Adversarial Networks (GANs)~\cite{goodfellow2014generative} has demonstrated strikingly high image synthesis quality, indistinguishable from real photographs~\cite{karras2019style,Karras2020stylegan2,karras2021alias}. However, contemporary generative approaches operate on 2D convolutional networks without modeling the underlying 3D scenes. Therefore 3D consistency cannot be strictly enforced when synthesizing head images under various poses. 

To generate 3D heads with diverse shapes and appearances, traditional approaches require a parametric textured mesh model~\cite{blanz1999morphable,FLAME:SiggraphAsia2017} learned from large 3D scan collections. However, the rendered images lack fine details and have limited perceptual quality and expressiveness. With the advent of differentiable rendering and neural implicit representation~\cite{mildenhall2020nerf,xie2022neural}, conditional generative models have been developed to generate more realistic 3D-aware face images~\cite{tewari2018self,tran2019towards,hong2022headnerf,zhuang2021mofanerf}. However, those approaches typically require multi-view image or 3D scan supervision, which are hard to acquire and have limited appearance distribution as those are usually captured in controlled environments. 



3D-aware generative models have recently seen rapid progress, fueled by the integration of implicit neural representation in 3D scene modeling and Generative Adversarial Networks (GANs) for image synthesis~\cite{schwarz2020graf,chan2021pi,niemeyer2021giraffe,chan2021efficient,epigraf,or2021stylesdf,xue2022giraffe}. Among them, the seminal 3D GAN, EG3D~\cite{chan2021efficient}, demonstrates striking quality in view-consistent image synthesis, trained only from in-the-wild single-view image collections. 
However, these 3D GAN approaches are still limited to synthesis in near-frontal views.

In this paper, we propose \emph{PanoHead}, a novel 3D-aware GAN for high-quality full 3D head synthesis in $360^\circ$ trained from only in-the-wild unstructured images. Our model can synthesize \emph{consistent 3D heads viewable from all angles}, which is desirable by many immersive interaction scenarios such as digital avatars and telepresence. To the best of our knowledge, our method is \emph{the first} 3D GAN approach to achieve full 3D head synthesis in $360^\circ$.

Extending 3D GAN frameworks such as EG3D~\cite{chan2021efficient} to full 3D head synthesis poses several significant technical challenges:
Firstly, many 3D GANs~\cite{chan2021efficient, or2021stylesdf} cannot separate foreground and background, inducing 2.5D head geometry. The background, formulated typically as a wall structure, is entangled with the generated head in 3D and therefore prohibits rendering from large poses. We introduce a \emph{foreground-aware tri-discriminator} that jointly learns the decomposition of the foreground head in 3D space by distilling the prior knowledge in 2D image segmentation. 

Secondly, while being compact and efficient, current hybrid 3D scene representations, like tri-plane~\cite{chan2021efficient}, introduce strong projection ambiguity for 360$^{\circ}$ camera poses, resulting in `mirrored face' on the back head. 
To address the issue, we present a novel 3D \emph{tri-grid volume representation} that disentangles the frontal features with the back head while maintaining the efficiency of tri-plane representations. 

Lastly, obtaining well-estimated camera extrinsics of in-the-wild back head images for 3D GANs training is extremely difficult. Moreover, an image alignment gap exists between these and frontal images with detectable facial landmarks. The alignment gap causes a noisy appearance and unappealing head geometry. Thus, we propose a novel \emph{two-stage alignment scheme} that robustly aligns images from any view consistently. This step decreases the learning difficulty of 3D GANs significantly. In particular, we propose a camera self-adaptation module that dynamically adjusts the positions of rendering cameras to accommodate the alignment drifts in the back head images.

Our framework substantially enhances the 3D GANs' capabilities to adapt to in-the-wild full head images from arbitrary views, as shown in Figure~\ref{fig:teaser}. The resulting 3D GAN not only generates high-fidelity 360$^{\circ}$ RGB images and geometry, but also achieves better quantitative metrics than state-of-the-art methods. With our model, we showcase compelling 3D full head reconstruction from a single monocular-view image, enabling easily accessible 3D portrait creation. 

In summary, our main contributions are as follows:

\begin{itemize}[leftmargin=10pt]
    \vspace{-0.05in}
    \item The first 3D GAN framework that enables view-consistent and high-fidelity full-head image synthesis with detailed geometry, renderable in $360^\circ$. We demonstrate our approach in high-quality monocular 3D head reconstruction from in-the-wild images.
    \vspace{-0.05in}
    \item A novel tri-grid formulation that balances efficiency and expressiveness in representing 3D $360^\circ$ head scenes.
    \vspace{-0.05in}
    \item A foreground-aware tri-discriminator that disentangles 3D foreground head modeling from 2D background synthesis.
    \item A novel two-stage image alignment scheme that adaptively accommodates  imperfect camera poses and misaligned image cropping, enabling training of 3D GANs from in-the-wild images with wide camera pose distribution.
\end{itemize}

\vspace{-0.1in}
\section{Related Work} \label{sec:related_work}

\begin{figure*}[t]
    \centering
    \includegraphics[width=0.95\textwidth]{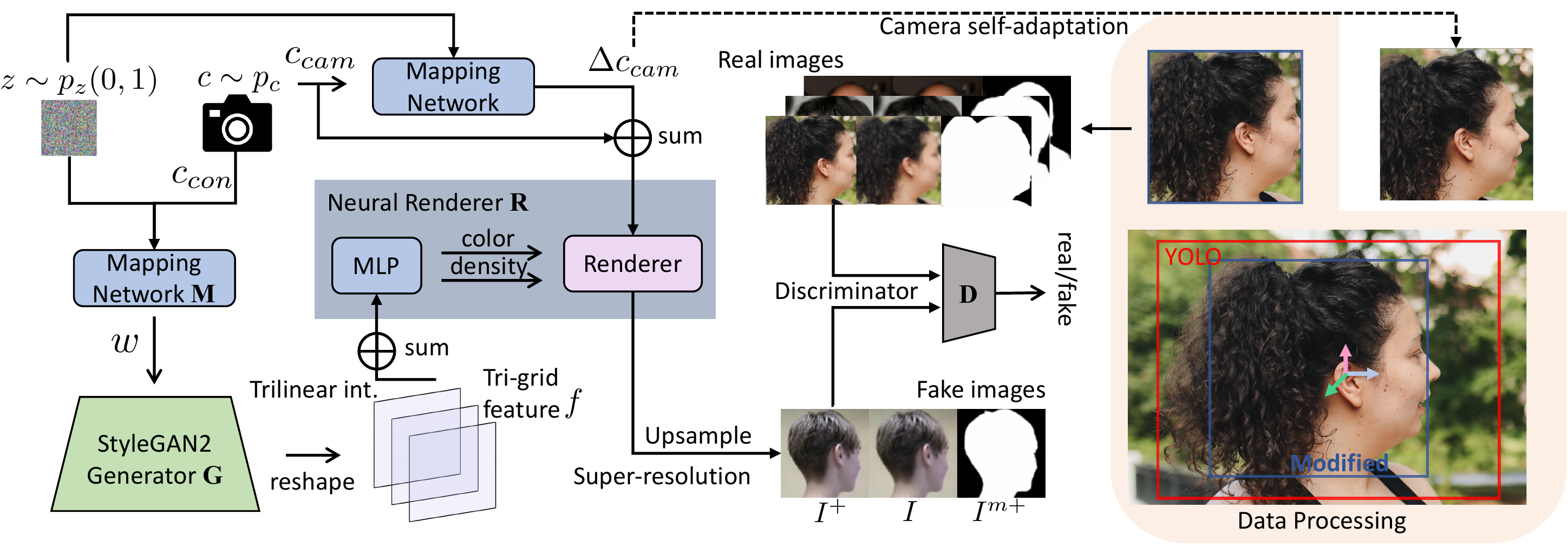}
    \caption{
     Our framework consists of three main components: a foreground-aware generator \bb{G}, discriminator \bb{D}, and a neural renderer \bb{R}. A mapping network first maps latent code $z$ and conditioned camera pose $c_{con}$ into the intermediate latent code $w$. The generator \bb{G} then takes $w$ to obtain the 3D tri-grid representation features $f$. With $f$ and rendering camera pose $c_{cam}$, the neural renderer \bb{R} synthesizes super-resolved image $I^{+}$, bilinear-upsampled image $I$, and super-resolved mask $I^{m+}$. Finally, the foreground-aware tri-discriminator \bb{D} critiques ($I^+$, $I$, $I^{m+}$) along with real images. The data processing pipeline is shown in the right side. The real images are cropped with modified YOLO bounding boxes yet they often differ at scale and location due to lacking accurate facial landmarks. With the camera self-adaptation scheme, the rendering camera pose $c_{cam}$ is able to correct itself to generate images with consistent scale and location.
    }
    \label{fig:overview}
    \vspace{-0.1in}
\end{figure*}

\paragraph{3D Head Representation and Rendering.}

To represent 3D heads with diverse shapes and appearances, a line of work has targeted parametric textured mesh representation, such as 3D Morphable Model (3DMM)~\cite{blanz1999morphable,paysan20093d,booth2018large,booth20163d} for faces and FLAME head model~\cite{FLAME:SiggraphAsia2017}, learned from 3D scans. However, these parametric representations do not model photo-realistic appearance and geometry beyond the front face or skull. The neural implicit functions~\cite{xie2022neural} have recently emerged as powerful continuous and differential representations of 3D scenes. Among them, Neural Radiance Field (NeRF)~\cite{mildenhall2020nerf,barron2021mip} has been widely adopted in digital head modeling~\cite{hong2022headnerf,park2021nerfies,gafni2021dynamic,guo2021adnerf,raj2021pva,szymanowicz2022photo} due to its superiority in modeling complex scene details and synthesizing multiview images with inherited 3D consistency. In contrast to optimizing a person-specific neural radiance field from multiview images or temporal videos, our approach builds a generative NeRF from unstructured 2D monocular images. Recently implicit-explicit hybrid 3D representation has been explored for better efficiency~\cite{chan2021efficient,devries2021unconstrained,martel2021acorn}. 
Among them, the tri-plane formulation proposed in EG3D~\cite{chan2021efficient} demonstrates a highly efficient 3D scene representation with high-quality view-consistent image synthesis. The tri-plane representation can scale efficiently with resolution, enabling greater detail for equal capacity. Our tri-grid representation transforms the tri-plane representation into a more expressive space for better feature embedding in unconditional 3D head synthesis. 

\paragraph{Single- or Few-view Supervised 3D GANs.}
Given the impressive progress of GANs on 2D image generation~\cite{goodfellow2014generative,karras2019style,Karras2020stylegan2,karras2021alias}, many studies have attempted to extend them to 3D-aware generation. These GANs aim to learn a generalizable 3D representation from 2D image a collections. For face synthesis,
Szabo~\etal~\cite{szabo2019unsupervised} first proposed using vertex position maps as the 3D representation to generate textured mesh outputs. Shi~\etal~\cite{shi2021lifting} proposed a self-supervised framework to convert 2D StyleGANs~\cite{karras2019style} into 3D generative models, although its generalizability is bounded by its base 2D StyleGAN. GRAF~\cite{schwarz2020graf} and pi-GAN~\cite{chan2021pi} are the first to integrate NeRF into 3D GANs. However, their performance is limited by the intense computation cost of forwarding and backwarding a complete NeRF. Many recent studies~\cite{deng2021gram,niemeyer2021giraffe,xue2022giraffe,gu2021stylenerf,or2021stylesdf,yariv2021volume,chan2021efficient,epigraf,gao2022get3d, enarf, schwarz2022voxgraf} have attempted to improve the efficiency and quality of such NeRF-based GANs. 
Specifically, EG3D~\cite{chan2021efficient}, which we build our work upon, introduces tri-plane representation that can leverage a 2D GAN backbone for generating efficient 3D representation and is shown outperforming other 3D representations~\cite{schwarz2022voxgraf}. 
Parallel to these works, another thread of studies~\cite{zhang2022avatargen,sun2022controllable,wu2022anifacegan,enarf} have been working on controllable 3D GANs that can manipulate the generated 3D faces or bodies. 


\section{Methodology} \label{sec:method}



\subsection{PanoHead Overview}
\label{sec:overview}
To synthesize realistic and view-consistent full head images, we build PanoHead upon a state-of-the-art 3D-aware GAN, \ie EG3D~\cite{chan2021efficient}, due to its efficiency and synthesis quality. 
Specifically, EG3D leverages StyleGAN2~\cite{Karras2020stylegan2} backbone to output a tri-plane representation that represents a 3D scene with three 2D feature planes. Given a desired camera pose $c_{cam}$, the tri-plane is decoded with a MLP network and volume rendered into a feature image, followed by a super-resolution module to synthesize a higher resolution RGB image $I^+$. Both the low and high resolution images are then jointly optimized by a dual discriminator $\bb{D}$.

In spite of EG3D's success in generating frontal faces, we found it to be a much more challenging task to adapt to 360$^\circ$ in-the-wild full head images for the following reasons: 1) foreground-background entanglement prohibit large pose rendering, 2) strong inductive bias from tri-plane representation causes mirroring face artifacts on the back head, and 3) noisy camera labels and inconsistent cropping of back head images. To 
address these problems, we introduce a background generator and a tri-discriminator for decoupling foreground and background (Section~\ref{sec:tridisc}), an efficient yet more expressive tri-grid representation while still being compatible with StyleGAN backbone (Section~\ref{sec:trigrid}), and a two-stage image alignment scheme with an self-adaptation module that dynamically adjusts rendering cameras during training (Section~\ref{sec:selftrans}). The overall pipeline for our model is illustrated in Figure~\ref{fig:overview}.

\subsection{Foreground-Aware Tri-Discrimination}
\label{sec:tridisc}

\begin{figure}[t]
    \centering
    \includegraphics[width=0.95\textwidth]{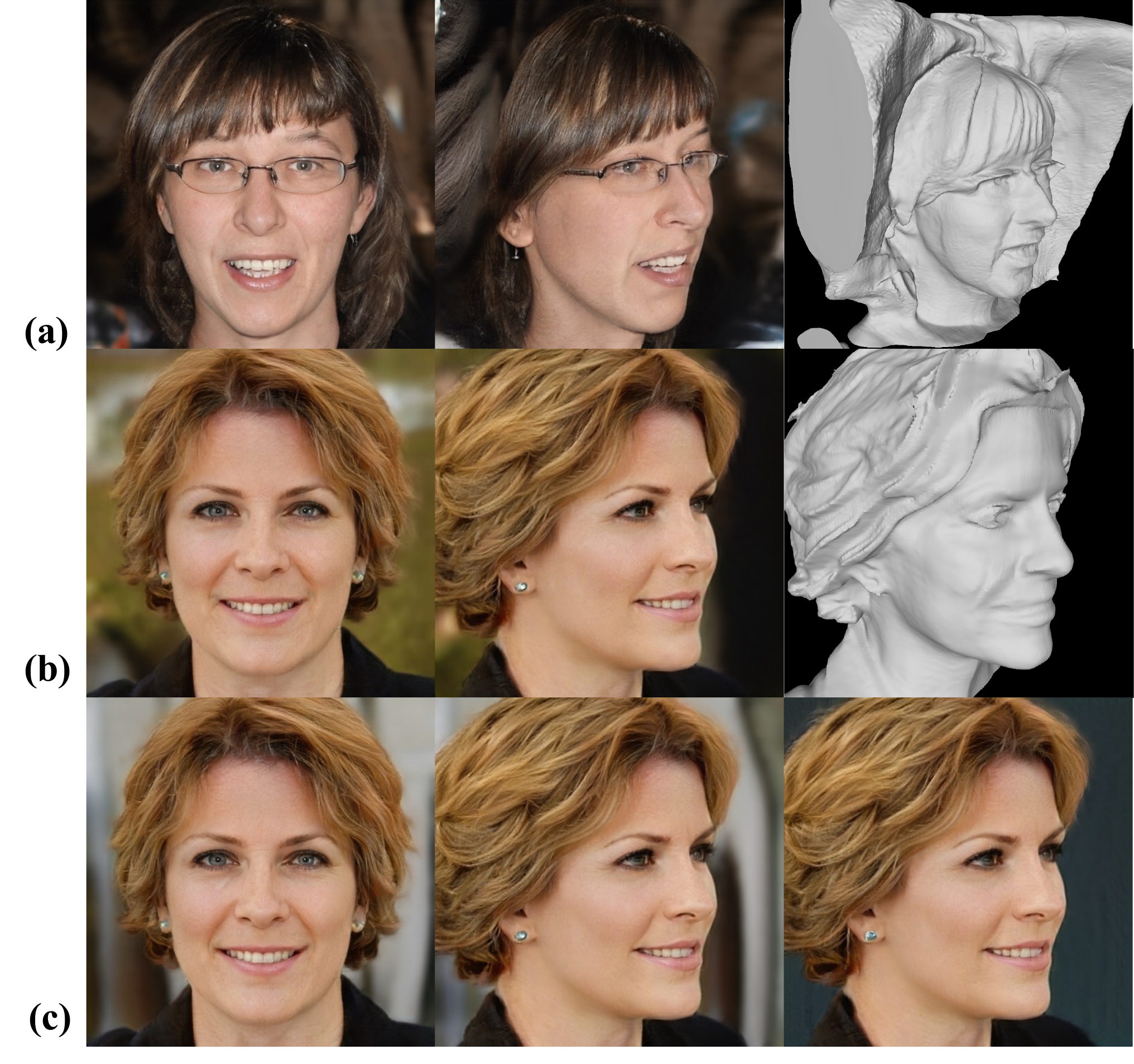}
    \caption{Geometry and RGB images from dual-discrimination (a) and foreground-aware tri-discrimination (b, c). EG3D (a) fails to decouple the background. PanoHead's tri-discrimination offers both background-free geometry (b) and background-switchable full head image synthesis (c).}
    \label{fig:tridisc_visual}
\end{figure}

A typical challenge of state-of-the-art 3D-aware GANs, like EG3D~\cite{chan2021efficient}, is the entangled foreground with the background of synthesized images. Regardless of the highly detailed geometry reconstruction, directly training the 3D GAN from in-the-wild RGB image collections, such as FFHQ~\cite{karras2019style}, results in a 2.5D face, as illustrated in Figure~\ref{fig:tridisc_visual} (a). Augmenting with image supervisions from the side and back of the head helps build up the full-head geometry with reasonable back head shapes. However, it does not solve the problem because the tri-plane representation itself is not designed to represent separated foreground and background.

To disentangle the foreground from the background, we first introduce an additional StyleGAN2 network~\cite{Karras2020stylegan2} to generate 2D backgrounds at the same resolution of raw feature image $I^r$. During volume rendering, the foreground mask $I^m$ can be obtained by:
\begin{gather}
    I^r(r) = \int_{0}^{\infty} w(t) f(r(t)) dt, \,\, I^m(r) = \int_{0}^{\infty} w(t) dt, \\
 w(t) = exp\bigl(-\int_{0}^{t} \sigma(r(s)) ds\bigr) 
    \sigma(r(t)),
\end{gather}
where $r(t)$ represents a ray emitted from the rendering camera center. The foreground mask is then used to compose a new low-resolution image $I^{gen}$:
\begin{equation}
    I^{gen} = (1 - I^m)I^{bg} + I^r,
\end{equation}
which is fed into the super-resolution module. Note that the computation cost of background generator is insignificant since its output has a much lower resolution than the tri-plane generator and super-resolution module.

Simply adding a background generator does not fully decouple it from the foreground since the generator tends to synthesize foreground content in the background. Thus, we propose a novel foreground-aware tri-discriminator to supervise the rendered foreground mask along with the RGB images. 
Specifically, the input of the tri-discriminator has 7 channels, composed with a bilinearly-upsampled RGB image $I$ , a super-resolved RGB image $I^+$ and single-channel upsampled foreground mask $I^{m+}$. The additional mask channel allows the 2D segmentation prior knowledge to be back-propagated into the density distribution of the neural radiance field. Our approach reduces the learning difficulty in shaping the 3D full head geometry from unstructured 2D images, enabling authentic geometry ((Figure~\ref{fig:tridisc_visual} (b))) and appearance synthesis of a full head composable with various backgrounds (Figure~\ref{fig:tridisc_visual} (c)). 
We note that in constrast from ENARF-GAN~\cite{enarf} that employs a single discriminator for RGB images composed of synthesized foreground and background images using a dual-generated mask, our tri-discriminator better ensures view-consistent high-resolution outputs.

\begin{figure}[b]
    \centering
    \includegraphics[width=0.95\textwidth]{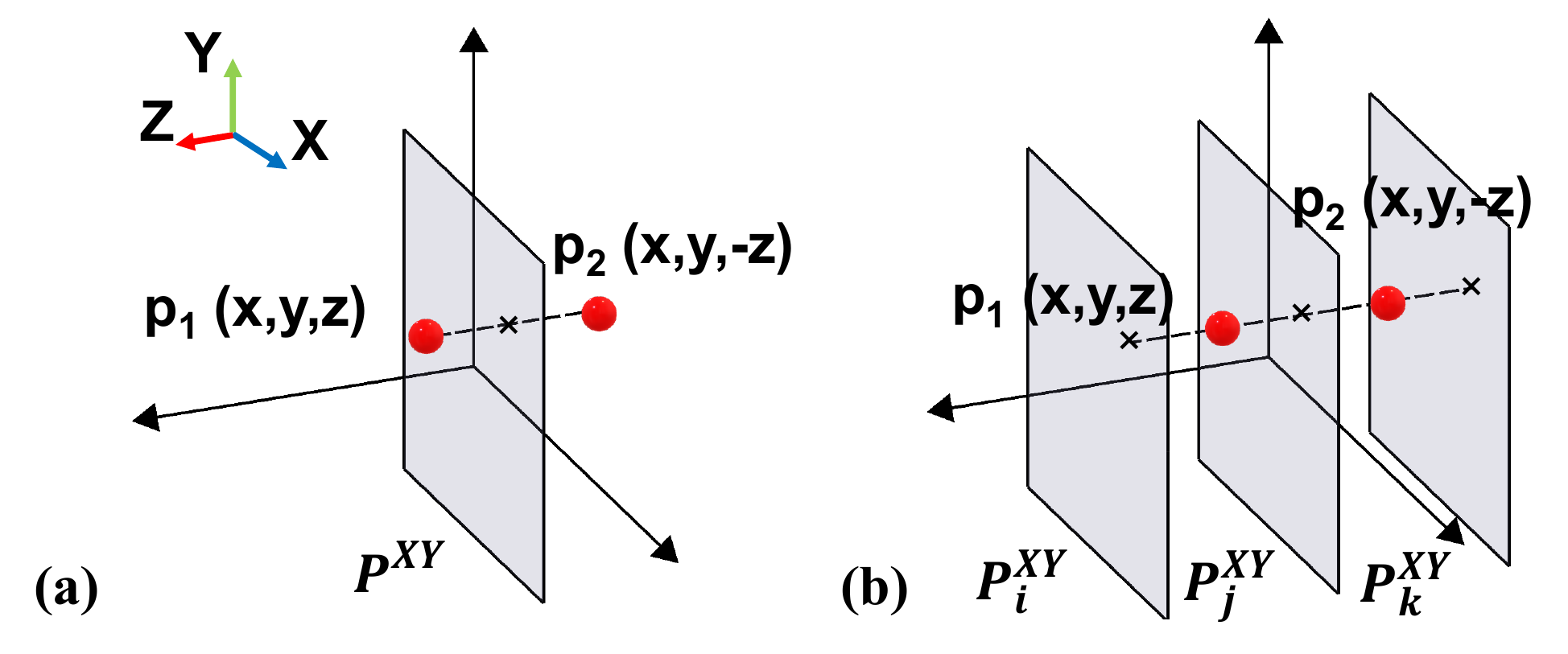}
    \caption{Comparison between tri-plane (a) and tri-grid (b) architecture in $Z$ axis. With tri-plane, two different points' projections share the feature from the plane $P^{XY}$, which introduces representation ambiguity. With tri-grid, the features for the above two points are trilinearly interpolated from two different planes, thus generating distinct features.} 
    \label{fig:trigrid_archi}
\end{figure}

\begin{figure}[t] 
    \centering
    \includegraphics[width=0.95\textwidth]{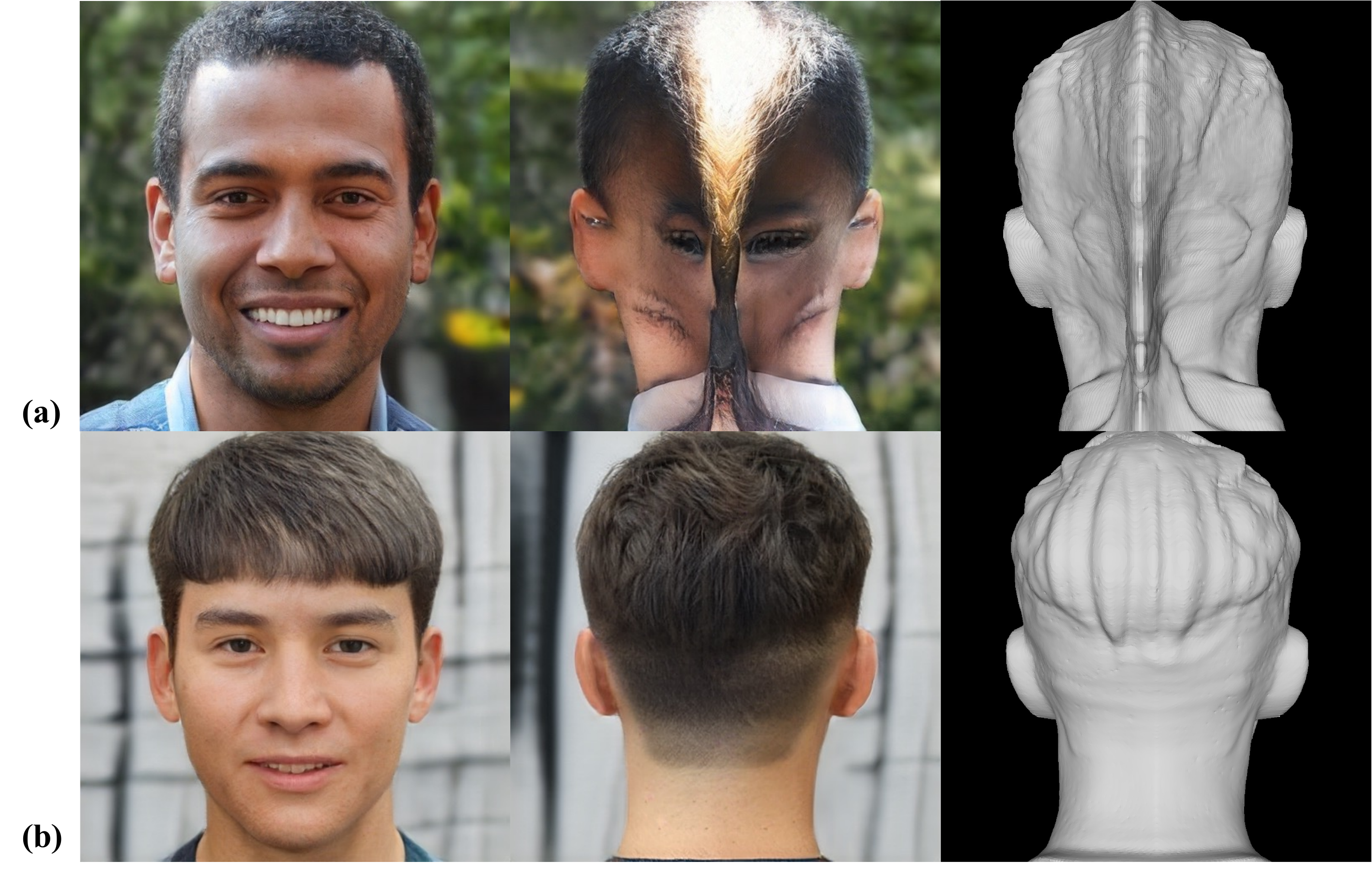}
    \caption{Images synthesis with tri-plane and tri-grid ($D=3$). Due to the projection ambiguity, tri-plane representation (a) can generate good-quality front face image yet with a `mirrored face' on back head, while our tri-grid representation synthesizes high-quality back head appearance and geometry (b).}
    \label{fig:trigrid_visual}
\end{figure}

\subsection{Feature Disentanglement in Tri-Grid}
\label{sec:trigrid}

The tri-plane representation, proposed in EG3D~\cite{chan2021efficient}, offers an efficient representation for 3D generation. The neural radiance density and appearance of a volume point are obtained by projecting its 3D coordinate over three axis-aligned orthogonal planes and decoding the sum of three bilinearly interpolated features with a tiny MLP. However, when synthesizing a full head in $360^\circ$, we observe tri-plane is limited in expressiveness and suffers from mirroring-face artifacts. 
The problem is even pronounced when the camera distribution of the training images is unbalanced. The root cause is the inductive bias originating from tri-plane projection, where one point on a 2D plane has to represent features of different 3D points. For example, a point on the front face and a point on the back hair will be projected to the same point on the $XY$ plane $P^{XY}$ (orthogonal to $Z$ axis), as illustrated in Figure~\ref{fig:trigrid_archi} (a). Although the other two planes should theoretically provide complementary information to alleviate this projection ambiguity, we found it not the case when there is less visual supervision from the back or when the structure of the back head is challenging to learn. The tri-planes are prone to borrow features from the front face to synthesize the back head, referred to as mirroring-face artifacts here (Figure~\ref{fig:trigrid_visual}(a)). 

To reduce the inductive bias of the tri-plane, we lift its formulation into a higher dimension by augmenting tri-plane with an additional depth dimension. 
We call this enriched version as a tri-grid. 
Instead of having three planes with a shape of $H \times W \times C$ with $H$ and $W$ being the spatial resolution and $C$ being the number of channel, each of our tri-grid has a shape of $D \times H \times W \times C$, where $D$ represents the depth. For instance, to represent spatial features on the $XY$ plane, tri-grid will have $D$ axis-aligned feature planes $P^{XY}_i, i = 1,\ldots, D$ uniformly distributed along the $Z$ axis. We query any 3D spatial point by projecting its coordinate onto each of the tri-grid, retrieving the corresponding feature vector by \emph{tri-linear interpolation}. As such, for two points sharing the same projected coordinates but with different depths, the corresponding feature would be likely to be interpolated from non-shared planes (Figure~\ref{fig:trigrid_archi} (b)). Our formulation disentangles the feature presentation of the front face and back head and therefore largely alleviates the mirroring-face artifacts (Figure~\ref{fig:trigrid_visual}). 

Similar to tri-plane in EG3D~\cite{chan2021efficient}, we can synthesize the tri-grid as $3\times D$ feature planes using the StyleGAN2 generator ~\cite{karras2019style}. That is, we increase the number of output channels of the original EG3D backbone by $D$ times. Thus, tri-plane can be regarded as a na\"ive case of our tri-grid representation with $D=1$. The depth $D$ of our tri-grid is tunable and larger $D$ offers more representation power at the cost of additional computation overhead. Empirically we find a small value of $D$ (\eg $D=3$) is sufficient in feature disentanglement while still maintaining its efficiency as a 3D scene representation. 

\begin{figure}[t]
    \centering
    \includegraphics[width=0.95\textwidth]{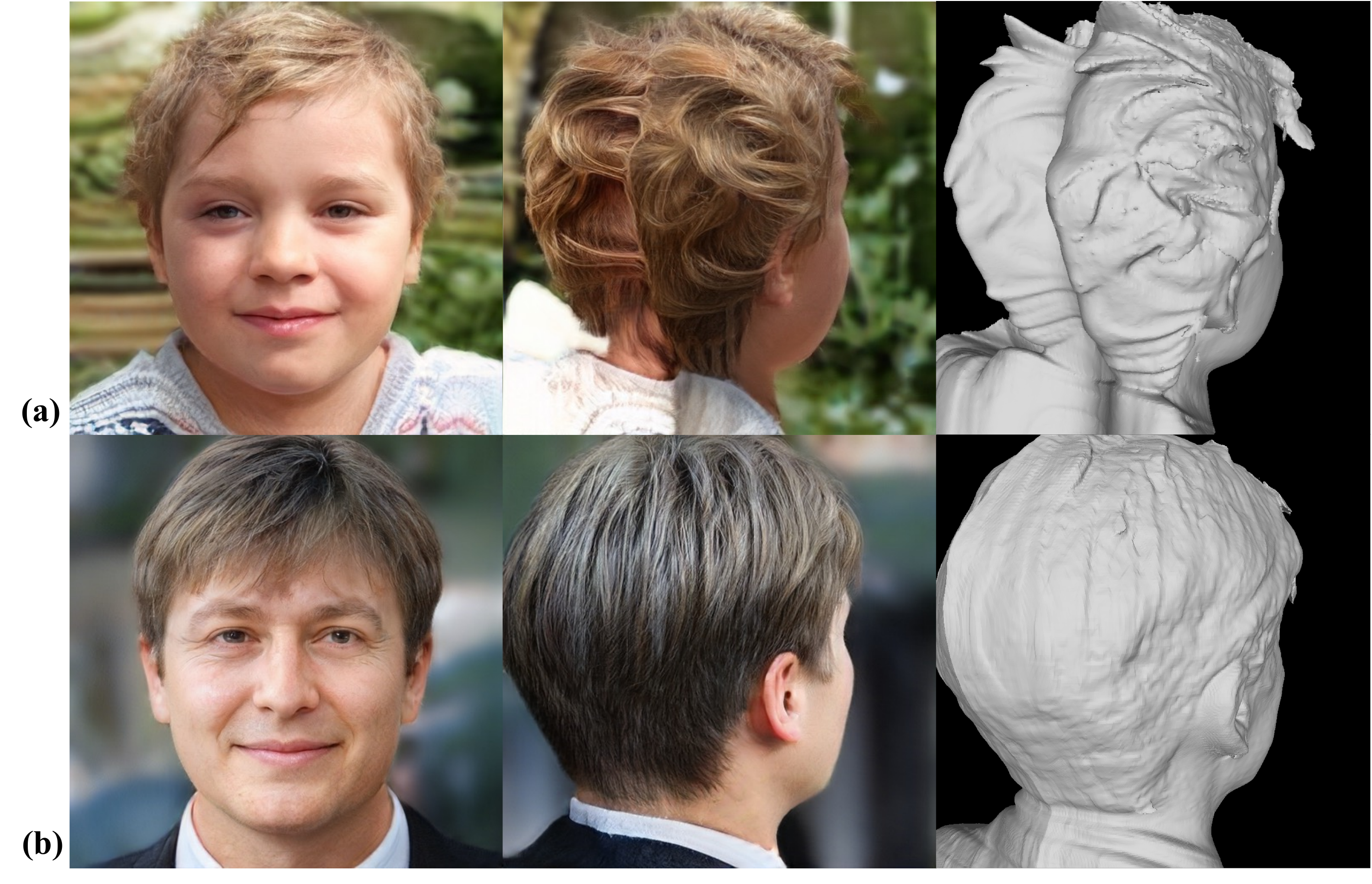}
    \caption{Image synthesized without (a) and with the camera self-adaptation scheme(b). Without it, the model generates misaligned back head images, leading to a defective dent in back head. }
    \label{fig:selftrans}
\end{figure}

\begin{figure*}[t]
    \centering
    \includegraphics[width=0.95\textwidth,trim=0 12 0 0,clip]{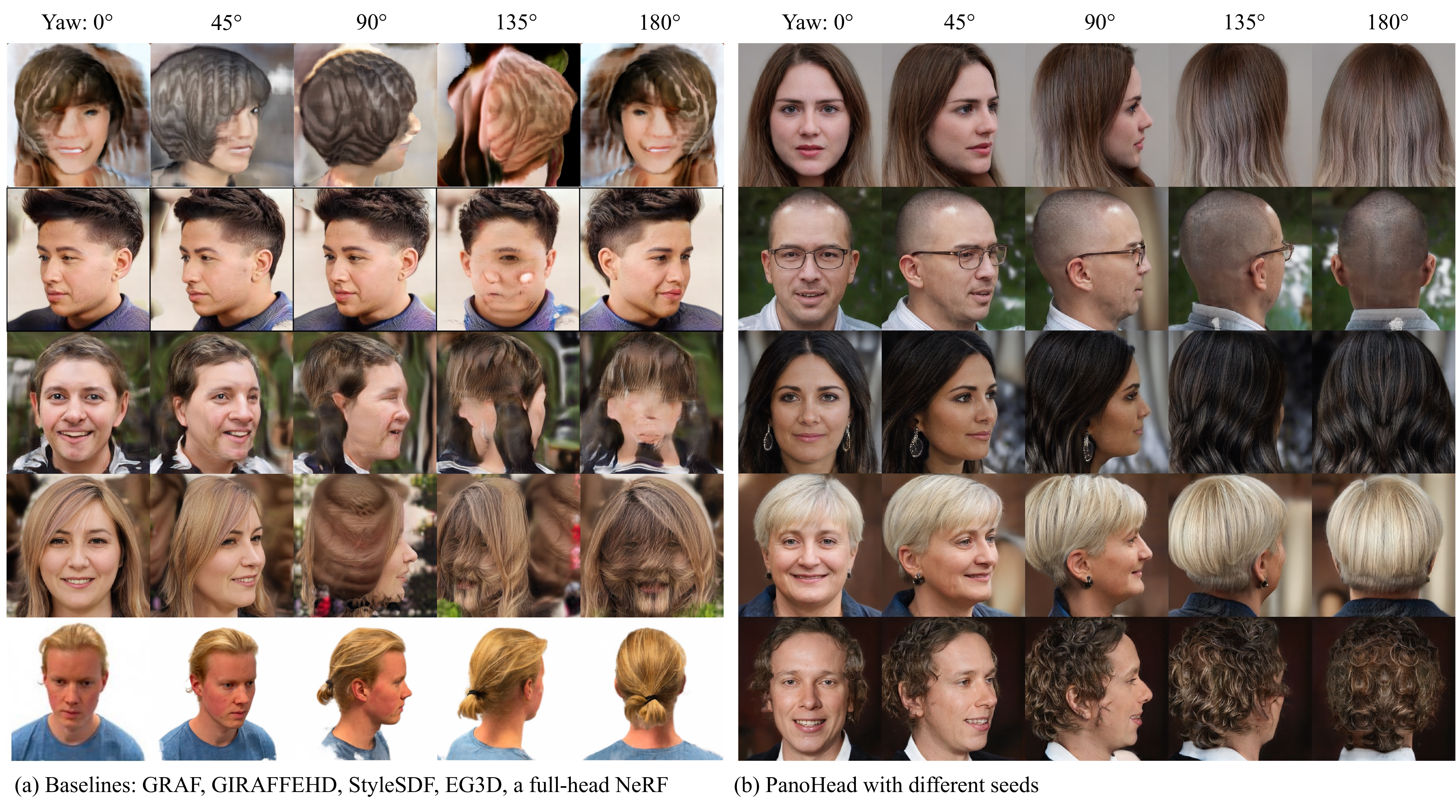}
    \caption{Qualitative comparison between GRAF~\cite{schwarz2020graf}, GIRAFFEHD~\cite{xue2022giraffe}, StyleSDF~\cite{or2021stylesdf}, EG3D~\cite{chan2021efficient}, multi-view supervised NeRF~\cite{szymanowicz2022photo} (different methods from top to bottom on left side), and our PanoHead (right). Except ~\cite{szymanowicz2022photo}, all models are trained on FFHQ-F. We render the results at a yaw angle of 0, 45, 90, 135, and 180$^{\circ}$. GRAF, GIRAFFEHD, and StyleSDF fail to model the correct camera distribution in latent space due to the unsupervised camera pose mechanism, thus not turning to the back. EG3D is able to rotate to the back with `mirroring face' artifacts and entangled background. Multi-view supervised NeRF is comparable to ours, however, it requires multi-view data of a single person and is not a generative model.
    }
    \label{fig:baselinecompare}
\end{figure*}


\subsection{Self-Adaptive Camera Alignment}
\label{sec:selftrans}
For adversarial training of our full head in $360^\circ,$  we need in-the-wild image exemplars from a much wider range of camera distribution than the mostly frontal distribution, as in FFHQ~\cite{karras2019style}. 
Although our 3D-aware GAN is only trained from widely-accessible 2D images, 
the key to the best quality training is accurate alignment of visual observations across images labeled with well-estimated camera parameters. While a good practice has been established for frontal face images cropping and alignment based on facial landmarks, it has never been studied in pre-processing large-pose images for GAN training. Both camera estimation and image cropping are no longer straightforward due to the lack of robust facial landmarks detection for images taken from the side and back.  

To resolve the aforementioned challenge, we propose a novel two-stage processing. In the first stage, for images with detectable facial landmarks, we still adopt the standard processing where the faces are scaled to a similar size and aligned at the center of the head using state-of-the-art face pose estimator 3DDFA~\cite{3DDFAv2}. For the rest of the images with large camera poses, we employ a head pose estimator WHENet~\cite{zhou2020whenet} that provides a roughly-estimated camera pose, and a human detector YOLO~\cite{ivavsic2019human} with a bounding box centered at the detected head. To crop the images at a consistent head scale and center, we apply both YOLO and 3DDFA on a batch of front-face images, from which we adjust the scale and translation of the head center of YOLO with constant offsets. 
This approach enables us to pre-process all head images with labeled camera parameters and in a consistent alignment to a large extent.

Due to the presence of various hairstyles, there is still inconsistency in the alignment of back head images, inducing significant learning difficulties for our network to interpret the complete head geometry and appearance (see Figure~\ref{fig:selftrans} (a)). We, therefore, propose a self-adaptive camera alignment scheme to fine-tune the transformation of volume rendering frustum for each training image. Specifically, our 3D-aware GAN associates each image with a latent code $z$ that embeds the 3D scene information of geometry and appearance, which can be synthesized at a view of $c_{cam}.$ $c_{cam}$ might not align well with the image content for our training images; so, it is hard for the 3D GAN to figure out a reasonable full head geometry. Therefore we co-learn a residual camera transformation $\Delta c_{cam}$ mapped from $(z, c_{cam})$ together with our adversarial training. The magnitude of $\Delta c_{cam}$ is regularized with a L$_2$ norm. Essentially, the network dynamically self-adapts the image alignment with refined correspondence across different visual observations. We note that this is only possible credited to the nature of 3D-aware GAN that can synthesize view-consistent images at various cameras. Our two-stage alignment enables 360-degree view-consistent head synthesis with authentic shape and appearance, learnable from diverse head images with widely distributed camera poses, styles, and structures.

\section{Experiments} \label{sec:exp}

\subsection{Datasets and Baselines}

We train and evaluate our framework on a balanced combination of FFHQ~\cite{karras2019style},  K-hairstyle dataset~\cite{kim2021k}, and an in-house large-pose head image collection. FFHQ contains $70K$ diverse high-resolution face images, yet mainly fall in the absolute yaw range from 0$^{\circ}$ to 60$^{\circ}$, assuming up-front camera pose corresponds to 0$^{\circ}$. 
We augment the FFHQ dataset with $4K$ back-head images from K-hairstyle dataset and $15K$ in-house large-pose images with diverse styles, ranging from 60$^{\circ}$ to 180$^{\circ}$. For brevity, we name this dataset combination as FFHQ-F. We refer to the supplementary paper for more dataset analysis and network training details.


We compare against state-of-the-art 3D-aware GANs including GRAF~\cite{schwarz2020graf}, EG3D~\cite{chan2021efficient}, StyleSDF~\cite{or2021stylesdf}, and GIRAFFEHD~\cite{xue2022giraffe}. All baselines are retrained from the same FFHQ-F dataset. We measure the quality of generated multiview images and geometry both quantitatively and qualitatively. 


\begin{table}[t]
\centering
\caption{Metrics comparison across all baselines.  For segmentation MSE, only GIRAFFEHD and PanoHead decouple the background and foreground. For ID score, GRAF's low-quality images lead to facial detection failure.}
\label{tab:quantitative}
\begin{adjustbox}{width=1\columnwidth,center}
\begin{tabular}{llllll}
\toprule
    & GRAF & GIRAFFEHD & StyleSDF & EG3D & Ours \\ \cmidrule{2-6} 
FID-all~$\downarrow$ & 68.2    & 37.3         & 78.5        & 6.2    & \textbf{5.4}    \\
MSE ($10^{-2}$)~$\downarrow$ & N/A    & 42.6         & N/A        & N/A    & \textbf{9.1}    \\
ID~$\uparrow$  & N/A    & 0.39         & 0.41        & \textbf{0.74}  & \textbf{0.74}    \\\bottomrule
\end{tabular}
\end{adjustbox}
\end{table}




\begin{table}[h]
\centering
\caption{Ablation studies on different components. +seg. means with foreground-aware tri-discrimination. +self-adpat. means with camera self-adaptation scheme. All are trained with FFHQ-F}
\label{tab:ablation}
\begin{adjustbox}{width=1\columnwidth,center}

\begin{tabular}{ccccc}
\toprule
        & EG3D  & \multicolumn{2}{c}{+seg.} & +seg.\&self-adapt. \\ \cmidrule{2-5} 
        &         &  tri-plane               & tri-grid                & tri-grid      \\ \cmidrule{3-5}
FID-back~$\downarrow$ & 50.4          & 44.1               & 44.0        & \textbf{40.9}               \\
FID-front~$\downarrow$ & 6.6        & \textbf{5.0}                & 5.5        & 5.4               \\
FID-all~$\downarrow$ & 6.2        & \textbf{5.2}                 & \textbf{5.2}        & 5.4               \\ \midrule
IS-back~$\uparrow$ & 4.3           & 3.9                & 4.2        & \textbf{4.4}               \\
IS-front~$\uparrow$ & 3.9           & \textbf{4.1}                & \textbf{4.1}        & \textbf{4.1}               \\
IS-all~$\uparrow$ & 3.8           & 4.0               & 4.0        & \textbf{4.1}               \\ \midrule
Runtime~$\downarrow$ & \textbf{1}      & 1.14$\times$                 & 1.26$\times$        & 1.28$\times$               \\ \bottomrule
\end{tabular}
\end{adjustbox}
\end{table}

\subsection{Qualitative Comparisons}

\noindent{\textbf{360$^\circ$ Image Synthesis.}}
 Figure~\ref{fig:baselinecompare} visually compares the image quality against the baselines, all trained with FFHQ-F, by synthesizing images from five different views, ranging the yaw angle from 0 to 180$^{\circ}.$  GRAF~\cite{schwarz2020graf} fails to synthesize compelling head images and its background is entangled with foreground head. StyleSDF~\cite{or2021stylesdf} and GIRAFFEHD~\cite{xue2022giraffe} are able to synthesize realistic frontal face images but in low perceptual quality when rendered from a larger camera pose. Without explicit reliance on camera labels, we suspect the above methods have difficulty in interpreting the 3D scene structures by themselves directly from images with 360$^{\circ}$ camera distribution.  
We observe that EG3D~\cite{chan2021efficient} is able to synthesize high-quality view-consistent frontal head images before rotating the view to the side or even the back. 
Mirroring face artifacts are clearly observable from the back, 
due to the tri-plane's projection ambiguity and the entangled fore-background. The method proposed in \cite{szymanowicz2022photo} builds personalized full-head NeRF at the extra cost of multi-view supervision. Regardless of its good quality images at all views, the model itself is not a generative model. In strong contrast, our model generates superior photo-realistic head images \textit{for all camera poses} while retaining multi-view consistency. It delivers photo-realism with fine details at diverse appearances, ranging from shaved head with glasses to long curly hairstyles. 
To better appreciate our multi-view full-head synthesis, please refer to our supplementary video for more comprehensive visual results.

\begin{figure}[t]
    \centering
    \includegraphics[width=1.\textwidth]{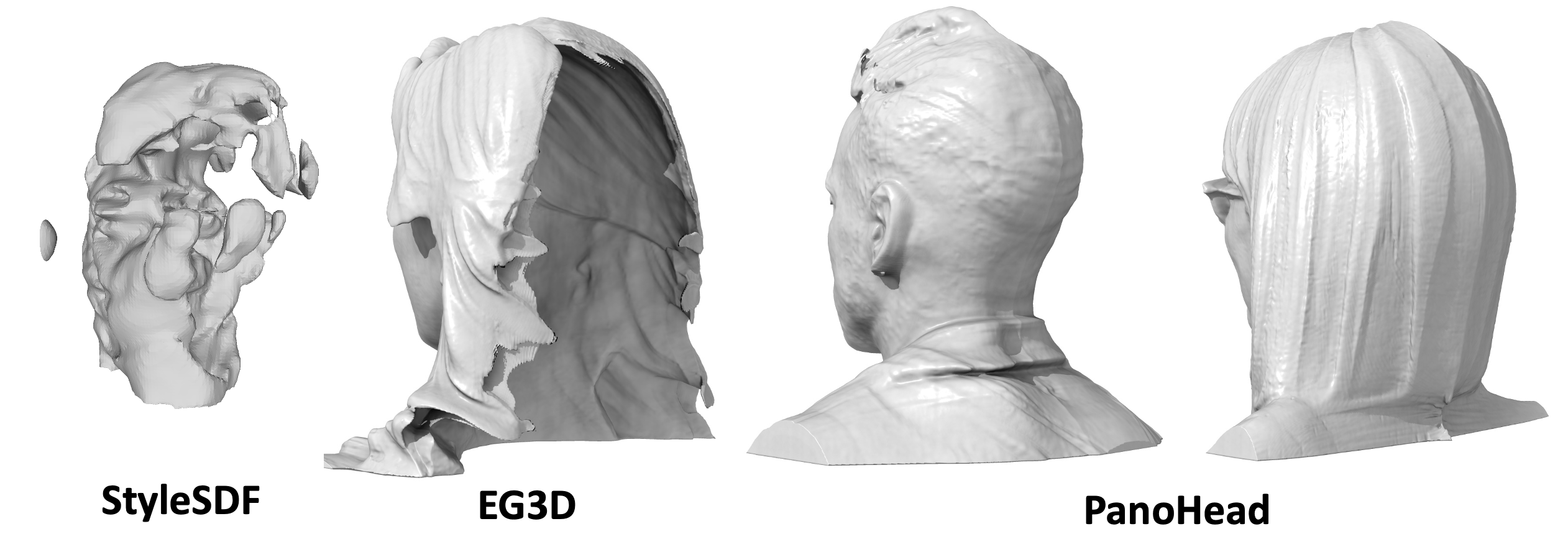}
    \caption{PanoHead achieves high-quality complete head geometry whereas StyleSDF~\cite{or2021stylesdf} and EG3D~\cite{chan2021efficient} produce 3D noises or hallowed heads.}
    \label{fig:geometry}
\end{figure}

\noindent{\textbf{Geometry Generation.}}
Figure~\ref{fig:geometry} compares the visual quality of the underlying 3D geometry extracted by running Marching Cubes algorithms~\cite{lorensen1987marching}. While StyleSDF~\cite{or2021stylesdf} generates decent appearances of the front face, the complete geometry of the head is noisy and broken. EG3D presents detailed geometry of front face and hair, but either with background concrete entangled (Figure~\ref{fig:tridisc_visual}(a)) or with a hollowed back head (Figure~\ref{fig:geometry}). 
In contrast, our model can consistently generate high-fidelity background-free 3D head geometry even with various hairstyles.

\subsection{Quantitative Results}

To quantify the visual quality, fidelity, and diversity of the generated images, we employ Frechet Inception Distance~(FID)~\cite{heusel2017gans} of 50K real and fake image samples.  We measure the multi-view consistency using the identity similarity score (ID) by calculating the average Adaface~\cite{kim2022adaface} cosine similarity score 
from paired synthesized face images rendered from different camera poses. Note that this metric can only be applied to those images with detected facial landmarks. We assess mean square error~(MSE) to calculate the accuracy of the generated segmentation against the mask obtained with DeepLabV3 ResNet101 network~\cite{chen2017rethinking}. Table~\ref{tab:quantitative} compares these metrics across all baselines and our method. We observe that our model outperforms other baselines consistently from all perspectives. Refer to supplemental material for metrics definition and implementation details.

To evaluate the image quality at different views, we employ FID and Inception Score~(IS)~\cite{salimans2016improved} for synthesized images with only back poses~($|yaw| \geq 90^\circ$), front poses~($|yaw| < 90^\circ$), and all camera poses. FID measures on the similarity and diversity of real and fake image distributions while IS focuses more on the image quality itself. 
Our GAN model follows EG3D for the main backbone, where the tri-plane generator is conditioned on a camera pose. We observe that such a design leads to biased image synthesis quality toward the conditioning camera pose. Specifically, when conditioning on the front view, our generator achieves inferior quality for synthesizing the head images from the back, and vice versa. However, when calculating FID-all, the conditioning camera is always the same as the rendering view. Therefore the generator could still achieve an excellent FID-all score even though the quality of generated heads might degenerate in unseen views. Hence, the original FID metrics (FID-all and FID-front) can hardly thoroughly reflect the overall generation quality of full heads in $360^{\circ}$.
To alleviate this issue, we propose FID-back, where we condition on the front view but synthesize the images from the back. It leads to higher FID scores but reflects the quality in $360^\circ$ image synthesis better.

We perform an ablation study on our method to quantitatively evaluate the efficacy of each individual component (Table~\ref{tab:ablation}). 
As shown in the second column, we notice a significant quality boost after adding the foreground-aware discrimination for all cases, compared with the original EG3D. That indicates the prior segmentation knowledge largely ease the network learning difficulty of 3D heads from in-the-wild image collections. Frontal face synthesis quality is comparable among all methods given the strong supervision from the large amount of well-aligned frontal images. However, for the back head, decoupling foreground and background largely improves the synthesis quality. In addition, changing tri-plane to tri-grid representation further enhances the image quality. With tri-discrimination, tri-grid, and camera self-adaption scheme altogether, PanoHead achieves the lowest FID-back and the highest IS for back head generation.
 As reflected in the row of run-time analysis, our novel component only introduces minor computation overhead, but with significant image synthesis quality improvements. 
 Note that the frontal image quality is superior to the back head, largely due to the significant learning difficulty in various hairstyles and unstructured back-head appearances. 



\subsection{Single-view GAN Inversion}
\label{sec:app}

\begin{figure}[t]
    \centering
    \includegraphics[width=1\textwidth]{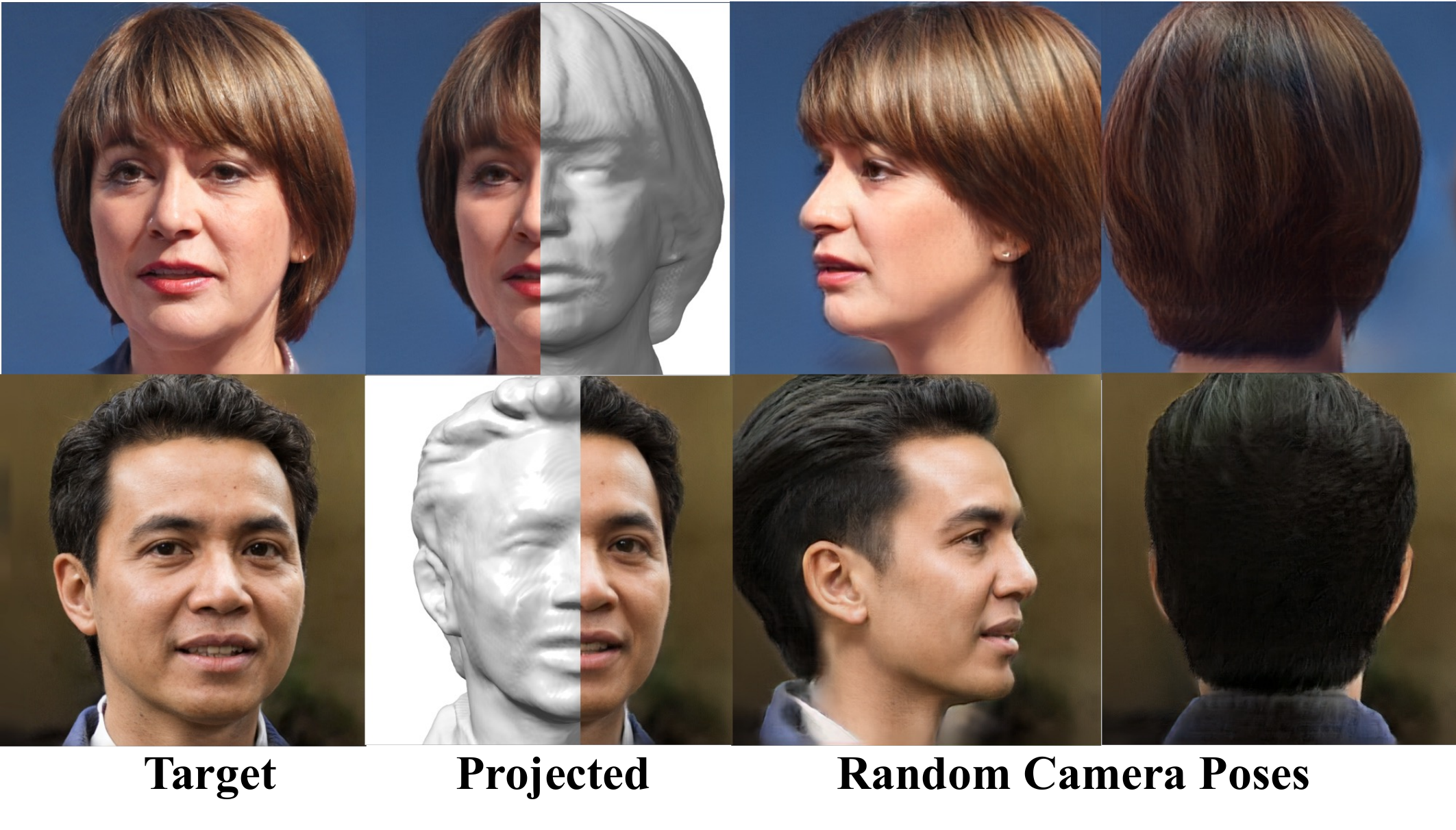}
    \caption{Single-view reconstruction from different camera poses. The first column shows the target images, second column projected RGB images and reconstructed 3D shapes using GAN inversion, last two columns rendered images from any given camera poses. }
    \label{fig:recon}
\end{figure}

Figure~\ref{fig:recon} demonstrates full-head reconstruction from a single-view portrait using PanoHead's generative latent space. 
To achieve that, we first perform an optimization to find the corresponding latent noise $z$ for the target image using pixel-wise L$_2$ loss and image-level  LPIPS loss~\cite{zhang2018unreasonable}. To further improve reconstruction quality, we perform pivotal tuning inversion~(PTI)~\cite{roich2021pivotal} to alter the generator parameters with a fixed optimized latent code $z.$ From a single-view target image, PanoHead not only reconstructs photo-realistic image and high-fidelity geometry but also enables novel-view synthesis in $360^\circ$, including large pose and back head. 

\section{Discussion} \label{sec:discussion}

\noindent{\textbf{Limitations and Future Work.}}
While PanoHead exhibits excellent images and shapes quality from $360^\circ$, it still contains minor artifacts, e.g. in the teeth area.  
Similar to the original EG3D, flickering texture issue is also noticeable in our model. Switching to StyleGAN3~\cite{Karras2021} as the backbone would help preserve high-frequency details. In practice, we also observe more noticeable flickering artifacts with a higher swapping probability of the conditional camera pose. We set this value to 70\% as opposed to 50\% in EG3D since we empirically find it enhances $360^\circ$ rendering quality but at the minor cost of flickering texture artifacts. 
Another observation is that it lacks finer high-frequency geometric details, e.g. hair tips. We leave it as future work to quantitatively evaluate our geometric quality such as using depth maps.
Finally, although PanoHead is able to generate diverse images in terms of gender, races, and appearances, reliance on training with only several datasets combination still makes it suffer from data bias, to some extent. In spite of our data collection effort, large-scale full-head annotated training image dataset is one of the most critical directions to facilitate full-head synthesis research. We anticipate such datasets can resolve some of the limitations aforementioned. 

\noindent{\textbf{Ethical considerations.}}
PanoHead is not specifically designed for any malicious uses, yet we do realize that the single-view portrait reconstruction could be manipulated, which might pose a social threat. We do not encourage the method being used for violating others' rights in any forms.
\section{Conclusion}
We propose PanoHead, the first 3D GAN framework that synthesizes view-consistent full head images with only single-view images. With our novel design in foreground-ware tri-discrimination, 3D tri-grid scene representation, and self-adaptive image alignment, PanoHead enables authentic multiview-consistent full-head image synthesis in $360^\circ$ and demonstrates compelling qualitative and quantitative results compared with state-of-the-art 3D GANs. Furthermore, we present 360-degree photo-realistic reconstruction
 with highly detailed geometry from single-view real portraits.  
 We believe the proposed method presents an interesting direction for 3D portraits creation, which sheds light on many potential downstream tasks.

{\small
\bibliographystyle{ieee_fullname}
\bibliography{egbib}
}

\end{document}